\documentclass[journal]{IEEEtran}
\usepackage{amssymb}
\usepackage{pifont,calc}
\usepackage{cite}

\usepackage{graphicx}
\ifCLASSINFOpdf
   \usepackage[pdftex]{graphicx}
\else
 \fi
\usepackage[cmex10]{amsmath}
\usepackage{array}

\usepackage{cases} % to combine equations together
\usepackage{bm} % to make xila zimu hei ti
\usepackage{subfigure}
\usepackage{mathrsfs}
\usepackage{url}
\usepackage[ruled]{algorithm2e}
\usepackage{arydshln}
\usepackage{multirow}
\usepackage{multicol}
\makeatletter

\newcommand{\Rmnum}[1]{\expandafter\@slowromancap\romannumeral #1@}
\makeatother
\ifCLASSINFOpdf
\else
\fi

\begin{document}
\title{Efficient Decremental Learning Algorithms  for Broad Learning System}

\author{Hufei~Zhu
\thanks{H. Zhu (e-mail: zhuhufei@aliyun.com).}}

\markboth{Journal of \LaTeX\ Class Files,~Vol.~14, No.~8, August~2015}%
{Shell \MakeLowercase{\textit{et al.}}: Bare Demo of IEEEtran.cls for IEEE Journals}

\maketitle

\begin{abstract}
 The decremental learning algorithms  are required in machine learning, to prune redundant nodes
 and remove obsolete inline training samples. In this paper, an efficient decremental learning algorithm to prune redundant nodes
 is deduced from the incremental learning algorithm $1$ proposed in  \cite{mybriefNodesOnBL} for added nodes,
 and   two decremental  learning algorithms to remove training samples are deduced from the two incremental learning algorithms proposed in \cite{mybriefInputsOnBL} for added inputs.
  The proposed decremental  learning algorithm
 for reduced  nodes utilizes
  the inverse Cholesky factor of the
   Hermitian matrix in the ridge inverse,  to update
  the output weights recursively,
as the incremental learning algorithm $1$ for added nodes in  \cite{mybriefNodesOnBL},
while that inverse Cholesky factor is updated with an unitary transformation.
The proposed decremental learning algorithm
  $1$
 for reduced  inputs
   updates
  the output weights  recursively with the inverse of
  the Hermitian matrix in the ridge inverse, and updates that inverse recursively,
   as the incremental learning algorithm $1$ for added inputs  in  \cite{mybriefInputsOnBL}.
   Moreover,
   the proposed decremental learning algorithm
  $2$
 for reduced  inputs
   updates
  the output weights  recursively with the inverse Cholesky factor of
  the Hermitian matrix in the ridge inverse, and updates that inverse Cholesky factor recursively by
  multiplying it
  with an upper-triangular intermediate matrix,
   as the incremental learning algorithm $2$ for added inputs  in  \cite{mybriefInputsOnBL}.
  In numerical experiments, all the proposed $3$ decremental  learning algorithms
 for reduced  nodes or  inputs  always achieve the testing accuracy of the standard ridge solution.
 When some nodes or  training samples are removed,
 the standard ridge solution by (\ref{W2AinvY989565ridge}) and (\ref{xAmnAmnTAmnIAmnT231413}) requires high computational complexity to  retrain the whole network from the beginning,
while
 the proposed decremental learning algorithms update
the output weights easily for reduced nodes or inputs without a retraining process.
\end{abstract}


\IEEEpeerreviewmaketitle

\section{Introduction}
%When Numda is big, my F will not be the same as BLS. When Numda is small, my ridge is also the same as BLS?
%all brief change into brief
%SLFN, training speed.
%
%BLS efficieient.
%
%reduced in my 3 papers. Ridge solutions.
%
%Decremental. Summary
  %, traditional train methods, e.g.,
In Single layer feedforward neural networks (SLFN) with the universal approximation capability~\cite{BL_Ref_18,BL_Ref_19,BL_Ref_20},
traditional  Gradient-descent-based learning algorithms~\cite{BL_Ref_22,BL_Ref_23} can be utilized,
%suffer from time-consuming training process.
%However, those Gradient-descent-based algorithms
which
suffer from the time-consuming training process.
%and the local minimum trap,
%% trap in a local minimum
%%converge slowly and may halt at a local minimum,
%while
%their generalization performance is sensitive to the
%training parameters,  e.g., learning rate.
To avoid the long training process,
 %Accordingly
  the random vector functional-link neural network (RVFLNN)
was proposed~\cite{BL_Ref_19}, which is a
universal approximation for continuous functions on
compact sets.
%to eliminate the drawback of long training process,
%which
%%RVFLNN
% offers the generalization capability in function
%approximation~\cite{BL_Ref_20}. It has been proven that RVFLNN is a
%universal approximation for continuous functions on
%compact sets.
%Based on the RVFLNN model,
The rapid incremental learning
 algorithm proposed in \cite{27_ref_BL_trans_paper}
updates the output weights of the RVFLNN easily for a new added
node or input, which can be applied to model  time-variety data with moderate size.
%When a new input is encountered or
%the increment of a new node is required,
%the dynamic algorithm in \cite{27_ref_BL_trans_paper} only computes the
%pseudoinverse of that added input or node,
%to update
%the output weights easily.
% for a new added node or input,
%by.
To deal with time-variety big data with
high dimension,
the scheme in \cite{27_ref_BL_trans_paper}
was improved
 into  Broad Learning
System (BLS)~\cite{BL_trans_paper}, which can
update
the output weights easily
 for any number of new added nodes or
inputs.
In \cite{BL_trans_paperApproximate}, the universal approximation capability of BLS was proved
 mathematically, and several BLS variants were proposed, which include cascade, recurrent, and broad-deep combination structures.
%  proof of
%is provided, and .
%In BLS~\cite{BL_trans_paper,BL_trans_paperApproximate},   the previous scheme~\cite{27_ref_BL_trans_paper} is improved in three aspects.
%Firstly, BLS
%%generates the mapped features from
%transforms
% the input data
% %to form
%  into
%  the feature nodes
%  to reduces the
%data dimensions.
% Secondly, BLS can update
%the output weights easily
% for any number of new added nodes or
%inputs, since it only requires one iteration to compute the
%pseudoinverse of those added nodes or inputs.
% Lastly,  to achieve a better generalization performance,

Efficient incremental learning BLS algorithms for new added
  nodes and inputs have been proposed in \cite{mybriefNodesOnBL}  and \cite{mybriefInputsOnBL}, respectively, to reduce the computational complexities
  of the original BLS algorithms in \cite{BL_trans_paper}. Moreover, the original BLS algorithms~\cite{BL_trans_paper}
%\cite{mybriefNodesOnBL, mybriefInputsOnBL}
% In the BLS discussed in ,  the incremental learning algorithms
  % compute
   %the output weights by
   utilize the ridge regression approximation of the generalized inverse, while  the BLS algorithms proposed in \cite{mybriefNodesOnBL, mybriefInputsOnBL}
   are based on the ridge inverse and the corresponding ridge solution~\cite{best_ridge_inv_paper213}.
   %Accordingly
   %which
  In \cite{BL_trans_paper}, the ridge parameter $\lambda \to 0$ is assumed in the ridge inverse~\cite{best_ridge_inv_paper213}
  to  approximate the generalized inverse, while in \cite{mybriefNodesOnBL, mybriefInputsOnBL},
  $\lambda$ can be any positive real number since
  the assumption of $\lambda \to 0$ is no longer required.
  %assure the ridge regression approximation of the generalized inverse.
  %    and no longer need
   %to  incremental learning algorithm for added nodes
   % Accordingly
 %   the assumption of
%     $\lambda \to 0$.
 %   (for
%  the generalized inverse with the ridge regression approximation in the existing BLS) is no longer required, and $\lambda$ can be any positive real number.
%  The efficient incremental learning algorithms for added inputs and for added nodes  have be proposed in BL paper, my 1 and 2.

 In machine learning, usually the decremental learning algorithms  are also required to prune redundant nodes~\cite{NodeDecremental1,NodeDecremental2,NodeDecremental3,NodeDecremental4,NodeDecremental5,NodeDecremental6}
 and remove obsolete inline training samples~\cite{InputDecremental1,InputDecremental2,InputDecremental3}.
 %Recently in \cite{my_brief1_on_BL},
% the inverse of a sum of matrices in \cite{InverseSumofMatrix8312} was utilized to improve the existing BLS on new added inputs in \cite{BL_trans_paper}, by
%  accelerating a step in the generalized inverse of a  row-partitioned matrix.
 Thus in this paper, we propose efficient decremental learning algorithms  to remove nodes and training samples, respectively.
 As the incremental learning algorithm $1$ for added nodes proposed in  \cite{mybriefNodesOnBL}, the proposed decremental  learning algorithm
 for reduced  nodes also computes the ridge solution (i.e,
  the output weights) from the inverse Cholesky factor of the
   Hermitian matrix in the ridge inverse. Moreover, the proposed decremental  learning algorithm
 for reduced  nodes updates the inverse Cholesky factor by an unitary transformation.
  %  and
% updates the inverse Cholesky factor  efficiently.
  On the other hand, as the incremental learning algorithms for added inputs proposed in  \cite{mybriefInputsOnBL}, the proposed two decremental  learning algorithms
 for reduced  inputs also
   compute
 the ridge solution from the inverse and the inverse Cholesky factor of
  the Hermitian matrix in the ridge inverse, respectively.
 %The proposed BLS algorithm $1$ utilizes the matrix inversion lemma~\cite{matrixInversionLemma} to
% %develop efficient algorithms to
%    update
%  the inverse  of
%  the Hermitian matrix.
% The proposed BLS algorithm $2$
% updates the upper-triangular  inverse Cholesky factor of  the Hermitian matrix,
% by multiplying the inverse Cholesky factor with an upper-triangular intermediate matrix, which is computed by
% a Cholesky factorization or an inverse Cholesky factorization.

  This paper is organized as follows. Section \Rmnum{2} introduces BLS and the efficient
  incremental learning algorithms proposed in \cite{mybriefNodesOnBL,mybriefInputsOnBL}.
 In Section \Rmnum{3}, we propose $1$ decremental learning algorithm for reduced nodes  and $2$ decremental learning algorithms
 for reduced inputs.
Then Section \Rmnum{4}  evaluates the  proposed decremental learning algorithms by numerical experiments.
Finally,
%we make conclusion
conclusions are given in Section \Rmnum{5}.

 \section{Broad Learning System and Incremental Learning Algorithms Proposed in \cite{mybriefNodesOnBL,mybriefInputsOnBL}}
% the original inputs are transferred and placed as
%※§ in feature nodes
%the mapped features are generated from the input data to form the feature nodes

 BLS transforms the original input ${{\mathbf{X}}}$
 into ``mapped features" in
 %to construct
  ``feature nodes" by some feature
mappings.  The
 feature nodes
 %which
 are then enhanced as the enhancement
nodes with random weights. All
 the feature nodes and the enhancement nodes form the expanded input matrix $\mathbf{A}$.
  The expanded input matrix $\mathbf{A}$
  can be denoted as
 %the
% $l \times k$
 %matrix
  $\mathbf{A}_{k}$ or
% $\mathbf{A}_{m\times n}\text{=}\mathbf{A}_{k}\text{=}\mathbf{A}_{{\bar{m}}}$ be , where the subscripts $n$ and $\bar{m}$ are the column number and the row number, respectively.
$\mathbf{A}_{{\bar{l}}}$, where the subscript $k$ denotes the column number and the total number of nodes,
 and the subscript  $\bar{l}$ denotes  the row number and the number of training samples.
% $\mathbf{A}_{m\times n}\text{=}\mathbf{A}_{k}\text{=}\mathbf{A}_{{\bar{m}}}$ be , where the subscripts $n$ and $\bar{m}$ are the column number and the row number, respectively.
 % In the BLS,  the original input data ${{\mathbf{X}}}$  is transferred into the mapped features in the feature nodes.
% Then the feature nodes are enhanced as the enhancement
%nodes. The expanded input matrix consists of all
% the feature nodes and the enhancement nodes, which can be written as
% %Denote
% the
% $l \times k$
% matrix
%% $\mathbf{A}_{m\times n}\text{=}\mathbf{A}_{k}\text{=}\mathbf{A}_{{\bar{m}}}$ be , where the subscripts $n$ and $\bar{m}$ are the column number and the row number, respectively.
%$\mathbf{A}_{{\bar{l}}}$, where the subscript  $\bar{l}$ denotes  the row number and the number of training samples, and the column number $k$ is equal to the node number.
The connections of all the feature
nodes and the enhancement nodes in $\mathbf{A}$ are fed into the output by
\begin{equation}\label{Y2AW65897}
{\mathbf{\hat{Y}}}=\mathbf{A}  {{\mathbf{W}}},
 \end{equation}%[]
 where ${{\mathbf{W}}}$ is the output weight matrix.
 %The least-square solution~\cite{27_ref_BL_trans_paper} of (\ref{Y2AW65897}) is
  The ridge solution~\cite{best_ridge_inv_paper213} of (\ref{Y2AW65897}) is
\begin{equation}\label{W2AinvY989565ridge}
{{\mathbf{W}}}=\mathbf{A}^{\dagger} \mathbf{Y},
\end{equation}
where ${{\bf{A}}^{\dagger }}$ is the ridge inverse~\cite{best_ridge_inv_paper213} of ${\bf{A}}$ that satisfies
\begin{equation}\label{xAmnAmnTAmnIAmnT231413}
{{\bf{A}}^{\dagger }}={{\left( {{\bf{A}}^{T}}{\bf{A}}+\lambda \mathbf{I} \right)}^{-1}}{{\bf{A}}^{T}}.
\end{equation}%[xAmnAmnTAmnIAmnT231413]

  The incremental learning algorithms were proposed in \cite{BL_trans_paper} to add
$q$  enhancement nodes and $p$ input training samples to the network, respectively.
%In this case, it
The $q$  enhancement nodes are added to
%which is equivalent to add $\mathbf{H}$ with $q$ columns to
 the input
matrix $\mathbf{A}_{k-q}^{{}}$ by
\begin{equation}\label{Anp2AnH954734}
\mathbf{A}_{k}^{{}}=\left[ \mathbf{A}_{k-q}^{{}}|\mathbf{H} \right],
 \end{equation}
 where $\mathbf{H}$ include $q$ columns.
 On the other hand, the additional $p$ input training samples can be denoted as ${{\mathbf{X}}_{a}}$,
%and
%The respectively increment of
and the incremental feature nodes and
enhancement nodes
corresponding to ${{\mathbf{X}}_{a}}$
can be denoted as  the
% $p \times k$
 matrix
 ${\bf{A}}_x$ with $p$ rows.
Accordingly the expanded input matrix $\mathbf{A}_{\bar l - \bar p}$ should be updated into
\begin{equation}\label{AxInputIncrease31232}
{\bf{A}}_{\bar l} = \left[ \begin{array}{l}
{\bf{A}}_{\bar l - \bar p}\\
{\bf{A}}_x^{}
\end{array} \right].
\end{equation}
%where the subscript  ${\bar l}$ denotes the row number and the total number of training samples.

The incremental learning algorithms in \cite{BL_trans_paper}  are based on the Greville's method~\cite{cite_general_inv_book}, which can only compute the generalized inverse of the partitioned matrices
(\ref{Anp2AnH954734}) and (\ref{AxInputIncrease31232}).
Correspondingly the ridge parameter
 $\lambda$  in (\ref{xAmnAmnTAmnIAmnT231413}) should be
  set to a very small positive real number,
%in \cite{BL_trans_paper},
e.g.,
${{10}^{-8}}$,
%$\lambda ={{10}^{-8}}$,
and then $\lambda \to 0$ can be assumed in (\ref{xAmnAmnTAmnIAmnT231413}) to approximate the generalized
inverse.
%assure that the ridge inverse always satisfies (\ref{AinvLimNumda0AAiA1221}).

The incremental learning algorithm $1$ proposed in \cite{mybriefNodesOnBL} and the two incremental learning algorithms proposed in \cite{mybriefInputsOnBL}  no longer need to assume $\lambda \to 0$,
which are all  based on the ridge inverse, and  always achieve the testing accuracy of the standard ridge solution in numerical experiments. However,
 usually
 the BLS algorithms in \cite{BL_trans_paper} achieve worse testing accuracy than the standard ridge solution in numerical experiments, when the assumption of $\lambda \to 0$
 is not satisfied (i.e., $\lambda$ is not very small).  %In the remainder of this section,
The rest of this section will give a brief introduction to the above-mentioned algorithms proposed in \cite{mybriefNodesOnBL,mybriefInputsOnBL}.

%As the incremental learning algorithm $1$ for added nodes proposed in  \cite{mybriefNodesOnBL}, the proposed decremental  learning algorithm
% for reduced  nodes also computes the ridge solution (i.e,
%  the output weights) from the inverse Cholesky factor of the
%   Hermitian matrix in the ridge inverse. Moreover, the proposed decremental  learning algorithm
% for reduced  nodes updates the inverse Cholesky factor by an unitary transformation.
%  %  and
%% updates the inverse Cholesky factor  efficiently.
%  On the other hand, as the incremental learning algorithms for added inputs proposed in  \cite{mybriefInputsOnBL}, the proposed two decremental  learning algorithms
% for reduced  inputs also
%   compute
% the ridge solution from the inverse and the inverse Cholesky factor of
%  the Hermitian matrix in the ridge inverse, respectively.

\subsection{Incremental Learning for Added Nodes}
%In this section, as an example, we
% only introduce
 The incremental learning algorithm $1$  proposed in  \cite{mybriefNodesOnBL} for added nodes
% , the proposed decremental  learning algorithm
% for reduced  nodes also
  computes the ridge solution (i.e,
  the output weights) from the inverse Cholesky factor of the
   Hermitian matrix in the ridge inverse,  and updates
   the inverse Cholesky factor efficiently by
 extending the  inverse Cholesky factorization in \cite{my_inv_chol_paper}.
%  the incremental learning algorithm to add
%$q$  new enhancement nodes to the network,
%%In this case, it
%which is equivalent to add $\mathbf{H}$ with $q$ columns to the input
%matrix $\mathbf{A}_{k-q}^{{}}$ by
%\begin{equation}\label{Anp2AnH954734}
%\mathbf{A}_{k}^{{}}=\left[ \mathbf{A}_{k-q}^{{}}|\mathbf{H} \right].
% \end{equation}

 Let
  \begin{equation}\label{R_define12321numda}
 {{\bf{R}}_{k}}={{\bf{A}}_{k}^T}{{\bf{A}}_{k}}+ \lambda {\bf{I}},
 \end{equation}
 and then (\ref{Anp2AnH954734}) can be substituted into  (\ref{R_define12321numda}) to obtain
 %can be represented as
  \begin{equation}\label{R_def_perhaps_No_R1numda}
  {{\bf{R}}_{k}}=\left[ \begin{matrix}
  {{\bf{R}}_{k-q}} & {{\bf{A}}_{k-q}^T}{{\bf{H}}}  \\
   {\bf{H}}^{T}{{\bf{A}}_{k-q}} & {\bf{H}}^{T}{{\bf{H}}}+\lambda {\bf{I}}  \\
\end{matrix} \right].
\end{equation}
The inverse Cholesky factor~\cite{my_inv_chol_paper} of ${{\bf{R}}_k}$
 %($i=k-q, k$)
%and ${{\bf{R}}_{k}}$
is the upper-triangular ${{\bf{F}}_k}$ that
%and ${{\bf{F}}_{k}}$, respectively,
satisfies
\begin{equation}\label{L_m_def12431before}
{{\bf{F}}_k}{{\bf{F}}_k^{T}}={{\bf{R}}_k^{-1}}=({{\bf{A}}_{k}^T}{{\bf{A}}_{k}}+ \lambda {\bf{I}})^{-1}.
\end{equation}
In \cite{mybriefNodesOnBL},  ${{\bf{F}}_k}$ is computed from ${{\bf{F}}_{k-q}}$ by
\begin{equation}\label{L_big_BLK_def1}{{\bf{F}}_{k}}=\left[ \begin{matrix}
   {{\bf{F}}_{k-q}} & \mathbf{T}  \\
   \mathbf{0} & \mathbf{G}  \\
\end{matrix} \right]
\end{equation}
where
\begin{subnumcases}{\label{ZF_def_L_2_items3ab}}
{{\mathbf{G}}}{{\mathbf{G}}^{T}}={\left( \begin{array}{l}
{{\bf{H}}^{T}{{\bf{H}}}+\lambda \mathbf{I}-{{\bf{H}}}^{T} \times}\\
{ {{\bf{A}}_{k-q}}{{\bf{F}}_{k-q}}{{\bf{F}}_{k-q}^{T}}{{\bf{A}}_{k-q}^T}{{\bf{H}}}}
\end{array} \right)^{ - 1}} &  \label{ZF_def_L_2_items3a}\\
 \mathbf{T}=-{{\bf{F}}_{k-q}}{{\bf{F}}_{k-q}^{T}}{{\bf{A}}_{k-q}^T}{{\bf{H}}}\mathbf{G}. & \label{ZF_def_L_2_items3b}
\end{subnumcases}
In (\ref{ZF_def_L_2_items3a}), the upper-triangular $\mathbf{G}$ is the inverse Cholesky factor of
${\bf{H}}^{T}{{\bf{H}}}+\lambda \mathbf{I}- {{\bf{H}}}^{T} {{\bf{A}}_{k-q}}{{\bf{F}}_{k-q}}{{\bf{F}}_{k-q}^{T}}{{\bf{A}}_{k-q}^T}{{\bf{H}}}$.

The output weight matrix
%Substitute (\ref{L_m_def12431before})  into
%(\ref{W2AinvY989565ridgeDecreaseNodes})
%to obtain
\begin{equation}\label{W2AinvToFDecreaseNodes134}
{{\mathbf{W}}_k}={{\bf{F}}_k}{{\bf{F}}_k^{T}}{{\bf{A}}_{k}^T} \mathbf{Y}
\end{equation}
is computed from ${{\bf{W}}_{k-q}}$ by
\begin{equation}\label{W2AMHEWNH3495}{ {\bf{W}}_{k} }=\left[ \begin{matrix}
   {{\bf{W}}_{k-q}}+\mathbf{T}{{\mathbf{G}}^{T}}\left( {\bf{H}}^{T}\mathbf{Y}-{{\bf{H}}^T}{{\bf{A}}_{k-q}}{{\bf{W}}_{k-q}} \right)  \\
   \mathbf{G}{{\mathbf{G}}^{T}}\left( {\bf{H}}^{T}\mathbf{Y}-{{\bf{H}}^T}{{\bf{A}}_{k-q}} {{\bf{W}}_{k-q}} \right)  \\
\end{matrix} \right].
\end{equation}

\subsection{Incremental Learning for Added Inputs}

%The BLS includes the incremental learning for the additional input training samples.
%When  encountering new input samples with the corresponding output labels,
%the modeled BLS can be remodeled in an
%incremental way without a complete retraining process.
%It updates the output weights incrementally, without retraining the whole network from the beginning.

The two incremental learning algorithms for added inputs proposed in \cite{mybriefInputsOnBL} compute
 the ridge solution (i.e., the output weights)
  from the inverse or the inverse Cholesky factor of
  the Hermitian matrix in the ridge inverse.
 The  algorithm $1$ proposed in \cite{mybriefInputsOnBL}
     updates
  the inverse  of
  the Hermitian matrix by the matrix inversion lemma~\cite{matrixInversionLemma},
  while the  algorithm $2$ proposed in \cite{mybriefInputsOnBL} updates
  %To  update
   the upper-triangular  inverse Cholesky factor of  the Hermitian matrix
 %the proposed BLS algorithm $2$
 by multiplying that inverse Cholesky factor with an upper-triangular intermediate matrix, which is computed by
 a Cholesky factorization or an inverse Cholesky factorization.

%Denote the additional input training samples  as ${{\mathbf{X}}_{a}}$.
%%and
%%The respectively increment of
% The incremental feature nodes and
%enhancement nodes
%corresponding to ${{\mathbf{X}}_{a}}$
%can be represented as  the
% $p \times k$
% matrix
% ${\bf{A}}_x$,
%and then the expanded input matrix $\mathbf{A}_{\bar l - \bar p}$ should be updated into
%\begin{equation}\label{AxInputIncrease31232}
%{\bf{A}}_{\bar l} = \left[ \begin{array}{l}
%{\bf{A}}_{\bar l - \bar p}\\
%{\bf{A}}_x^{}
%\end{array} \right],
%\end{equation}
%where the subscript  ${\bar l}$ denotes the row number and the total number of training samples.
Let ${{\mathbf{Y}}}$  and ${{\mathbf{Y}}_{a}}$ denote the output labels corresponding to
the input ${{\mathbf{X}}}$ and
the added input  ${{\mathbf{X}}_{a}}$, respectively, and write
\begin{equation}\label{}
{{\mathbf{Y}}_{\bar l}}=\left[ \begin{matrix}
   \mathbf{Y}  \\
   {{\mathbf{Y}}_{a}}  \\
\end{matrix} \right].
 \end{equation}%[]
Then
%Accordingly
 the  output weights (\ref{W2AinvY989565ridge})   can be written as
% ${\mathbf{W}}_{\bar l}$
%% should be into
%can be written as
\begin{equation}\label{xWbarMN2AbarYYa1341}
{\mathbf{W}}_{\bar l}={{\mathbf{Q}}_{\bar l}}{{\bf{A}}_{\bar l}^{T}}{{\mathbf{Y}}_{\bar l}},
 \end{equation}%[]
 % Write (\ref{xAmnAmnTAmnIAmnT231413}) as
%\begin{equation}\label{xA2QAt13413}
%{{\bf{A}}_{\bar l}^{\dagger }}={{\mathbf{Q}}_{\bar l}}{{\bf{A}}_{\bar l}^{T}}
%\end{equation}%[xA2QAt13413]
where
%the  $k \times k$ matrix
\begin{equation}\label{Qm1AAIdefine23213}
{{\mathbf{Q}}_{\bar l}}={{\left( {{\bf{A}}_{\bar l}^{T}}{\bf{A}}_{\bar l}+\lambda \mathbf{I} \right)}^{-1}}.
\end{equation}%[Qm1AAIdefine23213]

Substitute (\ref{AxInputIncrease31232}) into (\ref{Qm1AAIdefine23213}) to obtain
\begin{equation}\label{Q2AAIAtAinv4133467}
{{\mathbf{Q}}_{\bar l}}={{\left( \left( {{\bf{A}}_{\bar l - \bar p}^{T}}{\bf{A}}_{\bar l - \bar p}+\lambda \mathbf{I} \right)-\mathbf{A}_{x}^{T}(-\mathbf{A}_{x}^{{}}) \right)}^{-1}}.
\end{equation}%[Q2AAIAtAinv4133467]
Based on (\ref{Q2AAIAtAinv4133467}),
%Then
  the matrix inversion lemma ~\cite[equation (1a)]{matrixInversionLemma}
 and the inverse of a sum of matrices~\cite[equation (20)]{InverseSumofMatrix8312} were applied to
 %in myinputBLSpaper
 to deduce the algorithm $1$ in \cite{mybriefInputsOnBL}, i.e.,
\begin{subnumcases}{\label{yyyyxxxxx14223}}
p \le k: \left\{ \begin{array}{l}
{{\mathbf{B}}}= {{\mathbf{Q}}_{\bar l - \bar p} }\mathbf{A}_{x}^{T}{{(\mathbf{I}+\mathbf{A}_{x}^{{}}{{\mathbf{Q}}_{\bar l - \bar p} }\mathbf{A}_{x}^{T})}^{-1}} \\
{{\mathbf{Q}}_{\bar l}}={{\mathbf{Q}}_{\bar l - \bar p} }-{{\mathbf{B}}} \mathbf{A}_{x}^{{}}{{\mathbf{Q}}_{\bar l - \bar p} }
\end{array} \right.,  &  \label{DtAaaa1234yyyyxx}\\
p \ge k: \left\{ \begin{array}{l}
{{\mathbf{Q}}_{\bar l}}={{(\mathbf{I}+{{\mathbf{Q}}_{\bar l - \bar p} }\mathbf{A}_{x}^{T} \mathbf{A}_{x}^{{}})}^{-1}}{{\mathbf{Q}}_{\bar l - \bar p} } \\
{{\mathbf{B}}}= {{\mathbf{Q}}_{\bar l}} \mathbf{A}_{x}^{T}
\end{array} \right.,  & \label{CAxbbb14142yyyyxx}
\end{subnumcases}
and
\begin{equation}\label{xWmnIncreaseCompute86759}
{\mathbf{W}}_{\bar l}=\mathbf{W}_{\bar l - \bar p} + {{\mathbf{B}}}
  \left( {{\mathbf{Y}}_{a}}-\mathbf{A}_{x}^{{}}\mathbf{W}_{\bar l - \bar p}  \right).
 \end{equation}

The inverse Cholesky Factor~\cite{BL_trans_paper} of
 ${\bf{R}}_{\bar l }={{{{\bf{A}}_{\bar l }^{T}}{\bf{A}}_{\bar l }+\lambda \mathbf{I}}}$
  is the upper-triangular  ${{\mathbf{F}}_{\bar l } }$
 satisfying
\begin{equation}\label{Q2PiPiT9686954}
{\mathbf{F}}_{\bar l} {{\mathbf{F}}_{\bar l } ^{T}}={({{{{\bf{A}}_{\bar l }^{T}}{\bf{A}}_{\bar l }+\lambda \mathbf{I}}})^{-1}}={{\mathbf{Q}}_{\bar l } }.
\end{equation}%[Q2PiPiT9686954]
%the Cholesky Factor of ${{\mathbf{Q}}_{\bar l - \bar p} }$, i.e.,
Instead of updating the inverse
%${{\mathbf{Q}}_{\bar l - \bar p} }$ into
${{\mathbf{Q}}_{\bar l}}$ by (\ref{yyyyxxxxx14223}),
%From ${{\mathbf{F}}_{\bar l - \bar p} }$,
the algorithm $2$ proposed in \cite{mybriefInputsOnBL}
%we can also
 computes the inverse Cholesky factor
 %${{\mathbf{F}}_{\bar l - \bar p} }$ into
 ${{\mathbf{F}}_{\bar l}}$ from ${{\mathbf{F}}_{\bar l - \bar p} }$ by
\begin{equation}\label{Lbig2LLwave59056}
{{\mathbf{F}}_{\bar l}}={{\mathbf{F}}_{\bar l - \bar p} }\mathbf{V},
\end{equation}
where the upper-triangular $\mathbf{V}$ satisfies
\begin{subequations}{\label{Lwave2IKK40425both}}
 \begin{numcases}
{ {\mathbf{V}} {\mathbf{V}}^{T} = }
\mathbf{I}-{{\mathbf{S}}^{T}}{{(\mathbf{I}+\mathbf{S}{{\mathbf{S}}^{T}})}^{-1}}\mathbf{S}  \quad if\ p \le k &  \label{Lwave2IKK40425b} \\
{{(\mathbf{I}+{{\mathbf{S}}^{T}}\mathbf{S})}^{-1}}    \quad  \quad  \quad  \quad \  if\  p \ge k,  &  \label{Lwave2IKK40425a}
\end{numcases}
\end{subequations}
and  $\mathbf{S}$ in (\ref{Lwave2IKK40425both}) is computed by
\begin{equation}\label{K2AxLm94835}
\mathbf{S}=\mathbf{A}_{x}{{\mathbf{F}}_{\bar l - \bar p} }.
\end{equation}
Then the algorithm $2$ in \cite{mybriefInputsOnBL} computes ${\mathbf{W}}_{\bar l}$ from  $\mathbf{W}_{\bar l - \bar p}$
and  ${{\mathbf{F}}_{\bar l}}$ by
 %Moreover, W is computed from F by
\begin{equation}\label{xWmnIncreaseCompute4L9345}
{\mathbf{W}}_{\bar l}=\mathbf{W}_{\bar l - \bar p} + ({{\mathbf{F}}_{\bar l}}\mathbf{F}_{\bar l}^{T}\mathbf{A}_{x}^{{T}})
  \left( {{\mathbf{Y}}_{a}}-\mathbf{A}_{x}^{{}}\mathbf{W}_{\bar l - \bar p}  \right).
 \end{equation}

%Since both ${{\mathbf{F}}_{\bar l - \bar p} }$ and $\mathbf{V}$ are upper triangular,
% ${{\mathbf{F}}_{\bar l}}$ computed by  (\ref{Lbig2LLwave59056}) must be
%  upper triangular.
  %Notice that
  To reduce the computational complexity,
    a smaller inverse and inverse Cholesky factorization are chosen in (\ref{yyyyxxxxx14223}) and (\ref{Lwave2IKK40425both}),
  respectively,  according to the size of  the
 $p \times k$
 matrix $\mathbf{A}_{x}$.
Moreover, the upper-triangular inverse Cholesky factor in (\ref{Q2PiPiT9686954}) and (\ref{Lwave2IKK40425both})
%(\ref{Lwave2IKK40425a}) and
%(\ref{barL2IKKtinv4180842})
 can be computed by the inverse Choleksy factorization~\cite{my_inv_chol_paper},
%proposed in,
or by inverting and transposing the traditional lower-triangular Cholesky
factor~\cite{Matrix_Computations_book}.

\section{Proposed Decremental Learning Algorithms}

\subsection{Proposed Decremental Learning Algorithm for Reduced Nodes}

Assume the $\rho$ nodes corresponding to the columns $i_1,i_2,\cdots,i_{\rho}$ ($i_1<i_2<\cdots<i_{\rho}$) in $\mathbf{A}_{k}^{{}}$
needs to be removed. Then let us permute the columns $i_1,i_2,\cdots,i_{\rho}$ in $\mathbf{A}_{k}^{{}}$
to be the last $1^{st}, 2^{nd}, \cdots, {\rho}^{th}$ columns, respectively,  and the permuted $\mathbf{A}_{k}^{{}}$ can  be written as
% (\ref{Anp2AnH954734}),
\begin{equation}\label{Anp2AnH954734rho}
\mathbf{A}_{k}^{{}}=\left[ \mathbf{A}_{k-\rho}| \mathbf{A}_{\rho} \right],
 \end{equation}
 where $\mathbf{A}_{\rho}$ includes the $\rho$ columns to  be removed.
% Permute the columns in $\mathbf{A}_{k}^{{}}$ such that the $\rho$ nodes to  be removed
%are in the last $\rho$
Accordingly we need to
permute the rows $i_1,i_2,\cdots,i_{\rho}$ in $\mathbf{F}_{k}^{{}}$ and ${\bf{W}}_k$ to be  the last $1^{st}, 2^{nd}, \cdots, {\rho}^{th}$ rows, respectively,
as can be seen from (\ref{L_m_def12431before}) and (\ref{W2AinvY989565ridge}). The permuted $\mathbf{F}_{k}^{{}}$ can be block-triangularized by an unitary transformation ${\bf{\Sigma }}$, i.e.,
\begin{equation}\label{LbigBLKdef1Decrease}
 {{\bf{F}}_{k}} {\bf{\Sigma }} =\left[ \begin{matrix}
   {{\bf{F}}_{k-\rho}} & {\mathbf{T}}_{\rho}  \\
   \mathbf{0} & {\mathbf{G}}_{\rho}  \\
\end{matrix} \right],
\end{equation}
where ${\mathbf{T}}_{\rho}$ includes $\rho$ columns, and ${\mathbf{G}}_{\rho}$ is $\rho \times \rho$.
The sub-matrix ${{\bf{F}}_{k-\rho}}$ in (\ref{LbigBLKdef1Decrease}) is the ``square-root" matrix~\cite{ZhfSpLetter, my_inv_chol_paper}
 of ${{\bf{Q}}_{k-\rho}}={{\bf{R}}_{k-\rho}^{-1}}$ that satisfies ${{\bf{F}}_{k-\rho}} {{\bf{F}}_{k-\rho}^T} ={{\bf{Q}}_{k-\rho}}={{\bf{R}}_{k-\rho}^{-1}}$,
 %~\cite{my_inv_chol_paper},
as can be seen by comparing (\ref{LbigBLKdef1Decrease})  with (\ref{L_big_BLK_def1}).
From the permuted ${\bf{W}}_k$, the output weights  $\mathbf{W}_{k-\rho}^{{}}$ for the remaining $k-\rho$ nodes can be computed by
% it can easily be seen that.
 \begin{equation}\label{Wkrho2wkTGW834393}
\mathbf{W}_{k-\rho}^{{}}=\mathbf{W}_{k}^{1:(k-\rho),:}-\mathbf{T}_{\rho }^{{}}\mathbf{G}_{\rho }^{-1}\mathbf{W}_{k}^{(k-\rho +1):k,:},
 \end{equation}
%  From $\mathbf{T}_{\rho }^{{}}$ and $\mathbf{G}_{\rho }$
%in (\ref{LbigBLKdef1Decrease}),
where ${\bf{W}}_k^{i:j,:}$ denotes the sub-matrix in ${\bf{W}}_k$ from the $i^{th}$ row to the $j^{th}$ row. The derivation of
(\ref{Wkrho2wkTGW834393}) is in Appendix A.

The unitary transformation ${\bf{\Sigma }}$  in (\ref{LbigBLKdef1Decrease})  can be formed by a
sequence of Givens rotations~\cite{my_inv_chol_paper}.
%, by several Householder reflections~\cite{Matrix_Computations_book},
%or by a block Householder transformation~\cite{Block_househoulder}.
Assume $k=6$ and the rows $2$ and $4$ in $\mathbf{F}_{6}$ are permuted to be the last $1^{st}$ and  $2^{nd}$ rows, respectively.
The effect of the sequence of Givens rotations to triangularize the permuted $\mathbf{F}_{6}^{{}}$ can be shown as
%The $2^{nd}$ and $4^{th}$ rows are permuted to be the last and last $2^{nd}$ rows, respectively.
\begin{multline*}
\label{eqn_dbl_x}
\left|
\begin{smallmatrix}
 x & x & x & x & x & x \\
0& x & x & x & x & x \\
0&0& x & x & x & x \\
0&0&0& x & x & x \\
0&0&0&0& x & x \\
0&0&0&0&0& x
\end{smallmatrix}
          \right| \xrightarrow[]{} \left|
\begin{smallmatrix}
x & x & x & x & x & x \\
0&0& x & x & x & x \\
0&0&0&0& x & x \\
0&0&0&0&0& x \\
0&0&0& x & x & x \\
0& x & x & x & x & x
\end{smallmatrix}
          \right| \xrightarrow[{\bf{\Omega }}_{2,3}^6]{} \left|
\begin{smallmatrix}
 x & * & * & x & x & x \\
0&*& * & x & x & x \\
0&0&0&0& x & x \\
0&0&0&0&0& x \\
0&0&0& x & x & x \\
0& 0 &* & x & x & x
\end{smallmatrix}
          \right| \xrightarrow[{\bf{\Omega }}_{3,4}^6]{} \\
           \left|
\begin{smallmatrix}
  x & x & * &  * & x & x \\
0&x& * &  * & x & x \\
0&0&0&0& x & x \\
0&0&0&0&0& x \\
0&0&* & *  & x & x \\
0& 0 &0 & *  & x & x
\end{smallmatrix}
          \right| \xrightarrow[{\bf{\Omega }}_{4,5}^6]{} \left|
\begin{smallmatrix}
   x & x & x &  * & * & x \\
0&x& x &  * & * & x \\
0&0&0&*& * & x \\
0&0&0&0&0& x \\
0&0&x & *  & * & x \\
0& 0 &0 & 0  & * & x
\end{smallmatrix}
          \right| \xrightarrow[{\bf{\Omega }}_{5,6}^6]{} \left|
\begin{smallmatrix}
    x & x & x &  x & * & * \\
0&x& x &  x & * & * \\
0&0&0&x& * & * \\
0&0&0&0&*& * \\
0&0&x & x  & * &* \\
0& 0 &0 & 0  & 0 & *
\end{smallmatrix}
          \right| \xrightarrow[{\bf{\Omega }}_{3,4}^5]{}
          \\
           \left|
\begin{smallmatrix}
    x & x & * &  * & x & x \\
0&x& * &  * & x & x \\
0&0&*&*& x & x \\
0&0&0&0&x& x \\
0&0&0 & *  & x &x \\
0& 0 &0 & 0  & 0 & x
\end{smallmatrix}
          \right| \xrightarrow[{\bf{\Omega }}_{4,5}^5]{} \left|
\begin{smallmatrix}
     x & x & x &  * & * & x \\
0&x& x &  * & * & x \\
0&0&x&*& * & x \\
0&0&0&*&*& x \\
0&0&0 & 0  & * &x \\
0& 0 &0 & 0  & 0 & x
\end{smallmatrix}
          \right|,
\end{multline*}
where  ${\bf{\Omega }}_{m,n}^i$ is the Givens rotation that
 zeroes the $m^{th}$  entry in the $i^{th}$ row and
 rotates the $m^{th}$ and $n^{th}$ entries
 in each row of ${\bf{F }}_{k}$, and
 $*$ denotes the non-zero entries rotated by the current Givens rotation.
It can be seen that the sequence of Givens rotations
obtain the upper-triangular ${{\bf{F}}_{k-\rho}}$ and ${\mathbf{G}}_{\rho}$.

In (\ref{LbigBLKdef1Decrease}),
the unitary transformation ${\bf{\Sigma }}$  can also be formed by several Householder reflections~\cite{Matrix_Computations_book}
or  a block Householder transformation~\cite{Block_househoulder}.
When ${\bf{\Sigma }}$  is formed by a block Householder transformation, usually
%It can be seen that
  ${{\bf{F}}_{k-\rho}}$ and ${\mathbf{G}}_{\rho}$ in (\ref{LbigBLKdef1Decrease})
%obtained by the block Householder transformation
 are no longer upper-triangular.  Moreover, since there are many zeros in ${{\bf{F}}_{k}}$,  we can use a smaller block Householder transformation to reduce the computational complexity.
 For example, assume the permuted $\mathbf{F}_{6}^{{}}$ is
 \begin{displaymath}
  \mathbf{F}_{6}^{{}}=\left|
\begin{smallmatrix}
 x & x & x & x & x & x \\
0& x & x & x & x & x \\
0&0&0& x & x & x \\
0&0&0&0& x & x \\
0&0&0&0&0& x \\
0&0&0& x & x & x \\
0&0& x & x & x & x
\end{smallmatrix}
          \right|,
\end{displaymath}
and then we can set
${\bf{\Sigma }}=\left[ {\begin{array}{*{20}{c}}
{{{\bf{I}}_2}}&{\bf{0}}\\
{\bf{0}}&{\bf{\Theta }}
\end{array}} \right]$,  where ${{{\bf{I}}_2}}$ is the $2 \times 2$ identity matrix, and  ${\bf{\Theta }}$ is a block Householder transformation.
%Thus in step N3 the transformation $
%{\bf{\Sigma }} $ can be performed by only $(M-i)$ Givens rotations
%\cite{zhf_VTC08_6}, i.e.,
%\begin{equation}\label{equ:26GivensRotate321}
%{\bm \Sigma}^g_M = {\bf{\Omega }}_{i,i + 1}^i {\bf{\Omega }}_{i +
%1,i + 2}^i \cdots {\bf{\Omega }}_{M - 1,M}^i=\prod\limits_{j = i}^{M
%- 1} {{\bf{\Omega }}_{j,j + 1}^i },
%\end{equation}
%where the Givens rotation ${\bf{\Omega }}_{k,n}^i$
%% denotes  %matrix
%% that
% %to
% rotates the $k^{th}$ and $n^{th}$ entries %and the   entry
% in each row of ${\bf{F }}_{M}$,
% and zeroes the $k^{th}$  entry in the $i^{th}$ row.

%Substitute (\ref{L_m_def12431before})  into
%(\ref{W2AinvY989565ridgeDecreaseNodes})
%to obtain
%\begin{equation}\label{W2AinvToFDecreaseNodes134}
%{{\mathbf{W}}_i}={{\bf{F}}_i}{{\bf{F}}_i^{T}}{{\bf{A}}_{i}^T} \mathbf{Y}.
%\end{equation}

\subsection{Proposed Decremental Learning Algorithms for  Reduced  Inputs}

Permute the
rows in ${\bf{A}}_{\bar l}$
to put
%so that
 the  training samples to be removed into the sub-matrix
  ${\bf{A}}_{\bar \delta}$ with the last $\delta $ rows,
  i.e.,
\begin{equation}\label{AxInputDecrease31232}
{\bf{A}}_{\bar l} = \left[ \begin{array}{l}
{\bf{A}}_{\bar l - \bar \delta}\\
{\bf{A}}_{\bar \delta}
\end{array} \right],
\end{equation}
and permute the rows in ${{\mathbf{Y}}_{\bar l}}$ accordingly to obtain
%Accordingly
   \begin{equation}\label{Ypermuted7391742}
{{\mathbf{Y}}_{\bar l}}=\left[ \begin{matrix}
   {{\mathbf{Y}}_{\bar l - \bar \delta}}  \\
   {{\mathbf{Y}}_{\bar \delta}}  \\
\end{matrix} \right].
 \end{equation}

%As the algorithm $1$ in  myinputBLSpaper,
The proposed decremental learning algorithm  $1$  for reduced inputs
computes ${{\mathbf{Q}}_{\bar l - \bar \delta}}$
by
\begin{equation}\label{Qldelta2832DecreaseFinal}
{{\mathbf{Q}}_{\bar l - \bar \delta}}= {{\mathbf{\tilde B}}}{{\mathbf{Q}}_{\bar l} }
 \end{equation}
%in this paper
where
\begin{subequations}{\label{Bwave2832DecreaseFinal}}
 \begin{numcases}
{ \mathbf{\tilde B} = }
\mathbf{I} + {{\mathbf{Q}}_{\bar l} }{\bf{A}}_{\bar \delta}^{T}{{(\mathbf{I}-{\bf{A}}_{\bar \delta}^{{}}{{\mathbf{Q}}_{\bar l} }{\bf{A}}_{\bar \delta}^{T})}^{-1}}{\bf{A}}_{\bar \delta}^{{}} \quad if\ \delta \le k      &  \label{Bwave2832DecreaseFinala} \\
 {{(\mathbf{I}-{{\mathbf{Q}}_{\bar l} }{\bf{A}}_{\bar \delta}^{T} {\bf{A}}_{\bar \delta}^{{}})}^{-1}}    \quad  \quad   \quad   \quad  \quad \quad  \  if\  \delta \ge k,  &  \label{Bwave2832DecreaseFinalb}
\end{numcases}
\end{subequations}
and compute the output weights by
\begin{equation}\label{xWmnIncreaseCompute86759DecreaseFinal}
{\mathbf{W}}_{\bar l-\bar \delta}={{\mathbf{\tilde B}}}(\mathbf{W}_{\bar l }-{{\mathbf{Q}}_{\bar l} }{\bf{A}}_{\bar \delta}^{T} {{\mathbf{Y}}_{\bar \delta}}).
 \end{equation}
 The derivation of (\ref{Qldelta2832DecreaseFinal}),
(\ref{Bwave2832DecreaseFinal}) and
(\ref{xWmnIncreaseCompute86759DecreaseFinal}) is in Appendix B.

The
%proposed algorithm $2$
proposed decremental learning algorithm  $2$  for reduced inputs
updates ${{\mathbf{F}}_{\bar l} }$
 satisfying (\ref{Q2PiPiT9686954}) by
%The algorithm based on F is
\begin{equation}\label{Lbig2LLwave59056Decrease}
{{\mathbf{F}}_{\bar l - \bar \delta}}={{\mathbf{F}}_{\bar l} }\mathbf{V},
\end{equation}
and  the upper-triangular $\mathbf{V}$ in (\ref{Lbig2LLwave59056Decrease}) is computed by
\begin{subequations}{\label{Lwave2IKK40425bothDecrease}}
 \begin{numcases}
{ {\mathbf{V}} {\mathbf{V}}^{T} = }
\mathbf{I}+{{\mathbf{S}}^{T}}{{(\mathbf{I}-\mathbf{S}{{\mathbf{S}}^{T}})}^{-1}}\mathbf{S}  \quad if\ \delta \le k &  \label{Lwave2IKK40425bDecrease} \\
{{(\mathbf{I}-{{\mathbf{S}}^{T}}\mathbf{S})}^{-1}}    \quad  \quad  \quad  \quad \  if\  \delta \ge k,  &  \label{Lwave2IKK40425aDecrease}
\end{numcases}
\end{subequations}
%and  $\mathbf{S}$ in (\ref{Lwave2IKK40425bothDecrease}) is computed by
where
\begin{equation}\label{K2AxLm94835Decrease}
\mathbf{S}={\bf{A}}_{\bar \delta}{{\mathbf{F}}_{\bar l} }.
\end{equation}
Moreover,
the
proposed decremental learning algorithm  $2$  for reduced inputs
computes
the output weights ${\mathbf{W}}_{\bar l-\bar \delta}$
 %from ${{\mathbf{F}}_{\bar l - \bar \delta}}$
  by
\begin{equation}\label{xWmnIncreaseCompute4L9345Decrease}
{\mathbf{W}}_{\bar l-\bar \delta}={{\mathbf{F}}_{\bar l - \bar \delta}} {\mathbf{V}}^T {{\mathbf{F}}_{\bar l}^{-1}}\mathbf{W}_{\bar l }
- {{\mathbf{F}}_{\bar l - \bar \delta}}  {{\mathbf{F}}_{\bar l - \bar \delta}^T}   {\bf{A}}_{\bar \delta}^{{T}}{{\mathbf{Y}}_{\bar \delta}}.
 \end{equation}
 The derivation of (\ref{Lwave2IKK40425bothDecrease}) and (\ref{xWmnIncreaseCompute4L9345Decrease})  is also in Appendix B.

%{{\mathbf{F}}_{\bar l - \bar \delta}}={{\mathbf{F}}_{\bar l} }\mathbf{V}

To compute (\ref{Lwave2IKK40425bDecrease}), firstly compute the inverse Cholesky factor of $\mathbf{I}-\mathbf{S}{{\mathbf{S}}^{T}}$, i.e., the upper-triangular $\mathbf{\tilde{F}}$ satisfying
\begin{equation}\label{barL2IKKtinv4180842Decrease}
\mathbf{\tilde{F}}{{\mathbf{\tilde{F}}}^{T}}={{(\mathbf{I}-\mathbf{S}{{\mathbf{S}}^{T}})}^{-1}}.
\end{equation}
Then we need to compute the upper-triangular $\mathbf{V}$ satisfying
\begin{equation}\label{LLt2IKwaveKwaveT93224Decrease}
\mathbf{V}{{\mathbf{V}}^{T}}=\mathbf{I}+({{\mathbf{S}}^{T}}\mathbf{\tilde{F}}){({{\mathbf{S}}^{T}}\mathbf{\tilde{F}})^{T}},
\end{equation}  %{}    ()
where
the upper-triangular Cholesky factor
$\mathbf{V}$ is different from the traditional lower-triangular Cholesky factor~\cite{Matrix_Computations_book}.

\section{Numerical Experiments}

We compare the proposed decremental learning algorithms for BLS with
the standard ridge solution for BLS (by (\ref{W2AinvY989565ridge}) and (\ref{xAmnAmnTAmnIAmnT231413})),
%  and
% the existing BLS algorithm
  by the simulations
 on MATLAB software platform under a Microsoft-Windows Server with  $128$ GB of RAM.
We  give the experimental results on the
Modified National Institute of Standards and Technology (MNIST)
dataset~\cite{61_dataSet} with $60000$ training images and $10000$ testing images.
% In addition, the associated
%parameters Wei and 汕ei , for i = 1, . . . , n are drawn from the
%standard uniform distributions on the interval [?1, 1].
For the
enhancement nodes,
%to become the
%$j$-th group of enhancement nodes ${{\mathbf{H}}_{j}}$, where
the weights
    ${{\mathbf{W}}_{{{h}_{j}}}}$ and
   the biases
    ${{\mathbf{\beta }}_{{{h}_{j}}}}$ ($j=1,2,\cdots, m$)
    are drawn from the
standard uniform distributions on the interval $\left[ {\begin{array}{*{20}{c}}
{{\rm{ - }}1}&1
\end{array}} \right]$, and
 the sigmoid function is chosen.

 In Table  \Rmnum{1},
  we give the testing accuracy
of the standard ridge solution (by (\ref{W2AinvY989565ridge})  and
%xWbarMN2AbarYYa1341
(\ref{xAmnAmnTAmnIAmnT231413})) and the proposed decremental learning algorithm for reduced nodes,
which are abbreviated as  ``Standard" and ``Proposed", respectively.
  We set the initial network as $10 \times 10$
feature nodes and $11000$ enhancement nodes.
% The feature nodes are dynamically decreased
%from $100$  to $60$, and
Then the enhancement nodes are dynamically decreased
from $11000$ to  $7000$, and $\rho=1000$ enhancement nodes are removed in each update.
The snapshot results of each update are shown in Table \Rmnum{1}, where the ridge parameter $\lambda$
 is set
 to ${{10}^{-3}}$.
 %Table \Rmnum{2} and Table \Rmnum{3} show the testing accuracy
%of  the existing BLS algorithm, the proposed BLS algorithm $1$, the proposed BLS algorithm $2$ and
%the standard ridge solution (by (\ref{xWbarMN2AbarYYa1341})  and
%(\ref{xAmnAmnTAmnIAmnT231413})),
%which are abbreviated as  Existing, Alg. 1, Alg. 2 and Standard, respectively.
%
%We set
%%where
% the ridge parameter $\lambda$
% %is set
% to ${{10}^{-8}}$, ${{10}^{-6}}$, ${{10}^{-5}}$, ${{10}^{-4}}$, ${{10}^{-2}}$ and ${{10}^{-1}}$.
As observed from Table \Rmnum{1}, the proposed decremental
learning algorithm for reduced nodes always achieves the
testing accuracy of the standard ridge solution.

% However, the testing accuracy of the existing BLS algorithm is different from that
% of the standard ridge solution, and usually the difference becomes bigger when $\lambda$ is bigger.
% Moreover, when $\lambda$ is big (i.e., $\lambda \ge 10^{-4}$), usually
% the standard ridge solution achieves better testing accuracy than the existing BLS algorithm.

\begin{table}[!t]
\renewcommand{\arraystretch}{1.3}
\newcommand{\tabincell}[2]{\begin{tabular}{@{}#1@{}}#2\end{tabular}}
\caption{Snapshot Results of Testing Accuracy for the BLS Algorithms} \label{table_example} \centering
\begin{tabular}{|c||c c| c c|}
\hline
 \bfseries Number of       &\multicolumn{4}{c|}{{\bfseries   \tabincell{c}{$\lambda= {{10}^{-3}}$}}}    \\
 \bfseries Enhancement       &\multicolumn{2}{c|}{{\bfseries   \tabincell{c}{Testing Accuracy ($\% $)}}}  &\multicolumn{2}{c|}{{\bfseries   \tabincell{c}{Testing Accuracy ($\% $)}}} \\
 \bfseries  Nodes     &Standard   &Proposed    &Standard   &Proposed            \\
\cdashline{1-5}
 \bfseries 11000  & 97.23	&	97.23	&	96.95	&	96.95    \\
%\hline
 \bfseries 11000 $\to$ 10000   &  97.18	&	97.18	&	96.92	&	96.92     \\
%\hline
\bfseries 10000 $\to$ 9000 & 97.05	&	97.05	&	96.78	&	96.78     \\
%\hline
 \bfseries 9000 $\to$ 8000   & 96.99	&	96.99	&	96.74	&	96.74   \\
%\hline
 \bfseries 8000 $\to$ 7000   & 96.79	&	96.79	&	96.59	&	96.59    \\
\hline
\end{tabular}
\end{table}

We also simulate
 the decremental BLS on reduced inputs.
  %in Table  \Rmnum{2}.
   We set the network as $10 \times 10$
feature nodes and $5000$ enhancement nodes, and then the total node number  is $k=5100$.
In Table  \Rmnum{2},
 %and Table  \Rmnum{4},
 we train the initial network
under the first $l=60000$ training samples,
 %in the MNIST dataset,
and decrease $\delta=10000>k$
training samples %are added
 in each update, until only
$10000$ training samples are fed. On the other hand,
in Table  \Rmnum{3},
 %and Table  \Rmnum{5},
 we train the initial network
under the first $l=60000$ training samples,
 %in the MNIST dataset,
and decrease $\delta=1000<k$
training samples %are added
 in each update, until only
$55000$ training samples are fed.
The snapshot results of each update are shown
in
%the above-mentioned
 Tables  \Rmnum{2} and \Rmnum{3},
%, \Rmnum{4}
%and \Rmnum{5}
%show
where the ridge parameter $\lambda$
 is set
 to ${{10}^{-3}}$  and ${{10}^{-1}}$.  In Tables  \Rmnum{2} and \Rmnum{3},
 ``Standard" is the abbreviation of the standard ridge solution,
 while ``Alg. 1" and ``Alg. 2" are the abbreviations of  the proposed decremental learning algorithms $1$ and $2$ for
 reduced inputs, respectively.   As can be seen from Table \Rmnum{2}
 and Table \Rmnum{3},
 the proposed decremental learning algorithms $1$ and $2$ for reduced inputs always achieve the testing accuracy of the standard ridge solution.
  %the testing accuracy
%of the standard ridge solution (by (\ref{W2AinvY989565ridge})  and
%%xWbarMN2AbarYYa1341
%(\ref{xAmnAmnTAmnIAmnT231413})) and the proposed decremental learning algorithm for reduced nodes,
%which are abbreviated as  ``Standard" and ``Proposed", respectively.

%Table \Rmnum{2} and Table \Rmnum{3} show the testing accuracy
%of  the existing BLS algorithm, the proposed BLS algorithm $1$, the proposed BLS algorithm $2$ and
%the standard ridge solution (by (\ref{xWbarMN2AbarYYa1341})  and
%(\ref{xAmnAmnTAmnIAmnT231413})),
%which are abbreviated as  Existing, Alg. 1, Alg. 2 and Standard, respectively.
%We set
%%where
% the ridge parameter $\lambda$
% %is set
% to ${{10}^{-8}}$, ${{10}^{-6}}$, ${{10}^{-5}}$, ${{10}^{-4}}$, ${{10}^{-2}}$ and ${{10}^{-1}}$.
% As observed from Table \Rmnum{2} and Table \Rmnum{3},
% the proposed algorithms $1$ and $2$ both achieve the testing accuracy of the standard ridge solution.
% However, the testing accuracy of the existing BLS algorithm is different from that
% of the standard ridge solution, and usually the difference becomes bigger when $\lambda$ is bigger.
% Moreover, when $\lambda$ is big (i.e., $\lambda \ge 10^{-4}$), usually
% the standard ridge solution achieves better testing accuracy than the existing BLS algorithm.

\begin{table*}[!t]
\renewcommand{\arraystretch}{1.3}
\newcommand{\tabincell}[2]{\begin{tabular}{@{}#1@{}}#2\end{tabular}}
\caption{Snapshot Results of Testing Accuracy for $4$ BLS Algorithms with $p=10000>k=5100$} \label{table_example} \centering
\begin{tabular}{|c||c c c | c c c|c c c | c c c|}
\hline
\bfseries   Number of        &\multicolumn{6}{c|}{   {\bfseries   {$\lambda=10^{-3}$}} }   &\multicolumn{6}{c|}{ {\bfseries   {$\lambda=10^{-1}$}}  } \\
\bfseries    Input       &\multicolumn{3}{c|}{{\bfseries   {Training  Accuracy ($\% $)}} } &\multicolumn{3}{c|}{{\bfseries   {Testing  Accuracy ($\% $)}}}
&\multicolumn{3}{c|}{{\bfseries   {Training  Accuracy ($\% $)}}} &\multicolumn{3}{c|}{ {\bfseries   {Testing  Accuracy ($\% $)}} }  \\
\bfseries    Patterns  & Standard & Alg. 1     & Alg. 2 & Standard & Alg. 1     & Alg. 2 & Standard & Alg. 1     & Alg. 2 & Standard & Alg. 1     & Alg. 2  \\
\cdashline{1-13}
\bfseries 60000    &	96.87	&	96.87	&	96.87	&	96.77	&	96.77	&	96.77	&	92.58	&	92.58	&	92.58	&	93.05	&	93.05	&	93.05 \\
\hline
\bfseries  $\xrightarrow[\scriptscriptstyle{10000}]{}$ 50000    & 96.84	&	96.84	&	96.84	&	96.67	&	96.67	&	96.67	&	92.47	&	92.47	&	92.47	&	92.75	&	92.75	&	92.75	 \\
\hline
\bfseries  $\xrightarrow[\scriptscriptstyle{10000}]{}$ 40000     &	96.76	&	96.76	&	96.76	&	96.59	&	96.59	&	96.59	&	92.22	&	92.22	&	92.22	&	92.54	&	92.54	&	92.54 \\
\hline
\bfseries  $\xrightarrow[\scriptscriptstyle{10000}]{}$ 30000     &	96.58	&	96.58	&	96.58	&	96.34	&	96.34	&	96.34	&	91.82	&	91.82	&	91.82	&	92.23	&	92.23	&	92.23 \\
\hline
\bfseries  $\xrightarrow[\scriptscriptstyle{10000}]{}$ 20000      &	96.41	&	96.41	&	96.41	&	96.11	&	96.11	&	96.11	&	91.47	&	91.47	&	91.47	&	91.61	&	91.61	&	91.61 \\
\hline
\bfseries  $\xrightarrow[\scriptscriptstyle{10000}]{}$ 10000    &	96.19	&	96.19	&	96.19	&	95.62	&	95.62	&	95.62	&	90.54	&	90.54	&	90.54	&	90.71	&	90.71	&	90.71 \\
\hline
\end{tabular}
\end{table*}

\begin{table*}[!t]
\renewcommand{\arraystretch}{1.3}
\newcommand{\tabincell}[2]{\begin{tabular}{@{}#1@{}}#2\end{tabular}}
\caption{Snapshot Results of Testing Accuracy for $4$ BLS Algorithms with $p=1000<k=5100$} \label{table_example} \centering
\begin{tabular}{|c||c c c | c c c|c c c | c c c|}
\hline
\bfseries   Number of        &\multicolumn{6}{c|}{   {\bfseries   {$\lambda=10^{-3}$}} }   &\multicolumn{6}{c|}{ {\bfseries   {$\lambda=10^{-1}$}}  } \\
\bfseries    Input       &\multicolumn{3}{c|}{{\bfseries   {Training  Accuracy ($\% $)}} } &\multicolumn{3}{c|}{{\bfseries   {Testing  Accuracy ($\% $)}}}
&\multicolumn{3}{c|}{{\bfseries   {Training  Accuracy ($\% $)}}} &\multicolumn{3}{c|}{ {\bfseries   {Testing  Accuracy ($\% $)}} }  \\
\bfseries    Patterns  & Standard & Alg. 1     & Alg. 2 & Standard & Alg. 1     & Alg. 2 & Standard & Alg. 1     & Alg. 2 & Standard & Alg. 1     & Alg. 2  \\
\cdashline{1-13}
\bfseries 60000    &	96.34	&	96.34	&	96.34	&	96.21	&	96.21	&	96.21	&	92.83	&	92.83	&	92.83	&	93.24	&	93.24	&	93.24 \\
\hline
\bfseries  $\xrightarrow[\scriptscriptstyle{1000}]{}$ 59000    & 96.34	&	96.34	&	96.34	&	96.23	&	96.23	&	96.23	&	92.82	&	92.82	&	92.82	&	93.22	&	93.22	&	93.22	 \\
\hline
\bfseries  $\xrightarrow[\scriptscriptstyle{1000}]{}$ 58000     &	96.31	&	96.31	&	96.31	&	96.26	&	96.26	&	96.26	&	92.81	&	92.81	&	92.81	&	93.15	&	93.15	&	93.15 \\
\hline
\bfseries  $\xrightarrow[\scriptscriptstyle{1000}]{}$ 57000     &	96.31	&	96.31	&	96.31	&	96.22	&	96.22	&	96.22	&	92.81	&	92.81	&	92.81	&	93.11	&	93.11	&	93.11 \\
\hline
\bfseries  $\xrightarrow[\scriptscriptstyle{1000}]{}$ 56000     &	96.33	&	96.33	&	96.33	&	96.19	&	96.19	&	96.19	&	92.81	&	92.81	&	92.81	&	93.09	&	93.09	&	93.09 \\
\hline
\bfseries  $\xrightarrow[\scriptscriptstyle{1000}]{}$ 55000     &	96.32	&	96.32	&	96.32	&	96.22	&	96.22	&	96.22	&	92.78	&	92.78	&	92.78	&	93.07	&	93.07	&	93.07 \\
\hline
\end{tabular}
\end{table*}

 \section{Conclusion}

   The decremental learning algorithms  are required in machine learning, to prune redundant nodes~\cite{NodeDecremental1,NodeDecremental2,NodeDecremental3,NodeDecremental4,NodeDecremental5,NodeDecremental6}
 and remove obsolete inline training samples~\cite{InputDecremental1,InputDecremental2,InputDecremental3}. In this paper,
 an efficient decremental learning algorithm to prune redundant nodes
 is deduced from the incremental learning algorithm $1$ proposed in  \cite{mybriefNodesOnBL} for added nodes,
 and   two decremental  learning algorithms to
 remove training samples are deduced from the two incremental learning algorithms proposed in \cite{mybriefInputsOnBL} for added inputs. %respectively.

 %Both the incremental learning algorithm $1$ for added nodes in  \cite{mybriefNodesOnBL} and
%  the proposed decremental  learning algorithm
% for reduced  nodes
% utilize the inverse Cholesky factor of the
%   Hermitian matrix in the ridge inverse to update
%  the output weights recursively, by (\ref{W2AMHEWNH3495}) and
%(\ref{Wkrho2wkTGW834393}), respectively.

%In the incremental learning algorithm $1$ for added nodes in  \cite{mybriefNodesOnBL} and
%  the proposed decremental  learning algorithm
% for reduced  nodes,
%  the inverse Cholesky factor of the
%   Hermitian matrix in the ridge inverse is utilized to update
%  the output weights recursively, as can be seen from  (\ref{W2AMHEWNH3495}) and
%(\ref{Wkrho2wkTGW834393}),
%%Moreover, the proposed decremental  learning algorithm
%% for reduced  nodes updates the inverse Cholesky factor by an unitary transformation, as shown in (\ref{LbigBLKdef1Decrease}).
%Moreover, the proposed decremental  learning algorithm
% for reduced  nodes updates the inverse Cholesky factor with an unitary transformation by (\ref{LbigBLKdef1Decrease}).

  The proposed decremental  learning algorithm
 for reduced  nodes utilizes
  the inverse Cholesky factor of the
   Hermitian matrix in the ridge inverse,  to update
  the output weights recursively
  %, as can be seen from  (\ref{W2AMHEWNH3495}) and
  by
(\ref{Wkrho2wkTGW834393}),
as the incremental learning algorithm $1$ for added nodes in  \cite{mybriefNodesOnBL},
while that inverse Cholesky factor is updated with an unitary transformation by (\ref{LbigBLKdef1Decrease}).
The proposed decremental learning algorithm
  $1$
 for reduced  inputs
   updates
  the output weights  recursively with the inverse of
  the Hermitian matrix in the ridge inverse, and updates that inverse recursively,
   as the incremental learning algorithm $1$ for added inputs  in  \cite{mybriefInputsOnBL}.
   Moreover,
   the proposed decremental learning algorithm
  $2$
 for reduced  inputs
   updates
  the output weights  recursively with the inverse Cholesky factor of
  the Hermitian matrix in the ridge inverse, and updates that inverse Cholesky factor recursively by
  multiplying it
  %the inverse Cholesky factor
  with an upper-triangular intermediate matrix,
   as the incremental learning algorithm $2$ for added inputs  in  \cite{mybriefInputsOnBL}.

 % On the other hand, as the incremental learning algorithms $1$ and $2$ for added inputs  in  \cite{mybriefInputsOnBL}, the proposed decremental learning algorithms
%  $1$ and $2$
% for reduced  inputs
%   update
%  the output weights  recursively with the inverse and the inverse Cholesky factor of
%  the Hermitian matrix in the ridge inverse, respectively.
%In the proposed decremental learning algorithm
%  $1$, the inverse (of
%  the Hermitian matrix in the ridge inverse) is updated recursively, while in the proposed decremental learning algorithm
%  $2$, the inverse Cholesky factor (of
%  the Hermitian matrix in the ridge inverse) is updated by multiplying that inverse Cholesky factor
%  with an upper-triangular intermediate matrix.

 % On the other hand, as the incremental learning algorithms $1$ and $2$ for added inputs  in  \cite{mybriefInputsOnBL}, the proposed decremental learning algorithms
%  $1$ and $2$
% for reduced  inputs
%   update
%  the output weights  recursively with the inverse and the inverse Cholesky factor of
%  the Hermitian matrix in the ridge inverse, respectively.
%In the proposed decremental learning algorithm
%  $1$, the inverse (of
%  the Hermitian matrix in the ridge inverse) is updated recursively, while in the proposed decremental learning algorithm
%  $2$, the inverse Cholesky factor (of
%  the Hermitian matrix in the ridge inverse) is updated by multiplying that inverse Cholesky factor
%  with an upper-triangular intermediate matrix.

  In numerical experiments, all the proposed $3$ decremental  learning algorithms
 for reduced  nodes or  inputs  always achieve the testing accuracy of the standard ridge solution.
 When some nodes or  training samples are removed,
 the standard ridge solution by (\ref{W2AinvY989565ridge}) and (\ref{xAmnAmnTAmnIAmnT231413}) requires high computational complexity to  retrain the whole network from the beginning,
% With respect,
while
 the proposed decremental learning algorithms update
the output weights easily for reduced nodes or inputs without a retraining process.

\appendices

\section{Derivation of (\ref{Wkrho2wkTGW834393})}

From (\ref{W2AinvToFDecreaseNodes134}) we can deduce
${{\mathbf{W}}_k}={{\bf{F}}_k}{\bf{\Sigma }} {({\bf{F}}_k {\bf{\Sigma }})^{T}}{{\bf{A}}_{k}^T} \mathbf{Y}$, into which
we substitute (\ref{LbigBLKdef1Decrease}) and (\ref{Anp2AnH954734rho})
%Substitute
%(\ref{LbigBLKdef1Decrease})
%into (\ref{L_m_def12431before}), and then substitute (\ref{L_m_def12431before}) and (\ref{Anp2AnH954734}) into
%(\ref{W2AinvY989565ridgeDecreaseNodes})
to obtain
\begin{equation}\label{W2AinvY989565ridgeDecreaseNodesU1}
{{\mathbf{W}}_k}=\left[ \begin{matrix}
   {{\bf{F}}_{k-\rho}} & {\mathbf{T}}_{\rho}  \\
   \mathbf{0} & {\mathbf{G}}_{\rho}  \\
\end{matrix} \right]\left[ \begin{matrix}
   {{\bf{F}}_{k-\rho}} & {\mathbf{T}}_{\rho}  \\
   \mathbf{0} & {\mathbf{G}}_{\rho}  \\
\end{matrix} \right]^T {\left[ \mathbf{A}_{k-\rho}| \mathbf{A}_{\rho} \right]^T} \mathbf{Y},
\end{equation}
i.e.,
   \begin{multline*}\label{}
\mathbf{W}_{k}^{{}}= \\
\left[ \begin{matrix}
   \mathbf{F}_{k-\rho }^{{}}\mathbf{F}_{k-\rho }^{T}\mathbf{A}_{k-\rho }^{T}\mathbf{Y}+\mathbf{T}_{\rho }^{{}}(\mathbf{T}_{\rho }^{T}\mathbf{A}_{k-\rho }^{T}\mathbf{Y}+\mathbf{G}_{\rho }^{T}\mathbf{A}_{\rho }^{T}\mathbf{Y})  \\
   \mathbf{G}_{\rho }^{{}}(\mathbf{T}_{\rho }^{T}\mathbf{A}_{k-\rho }^{T}\mathbf{Y}+\mathbf{G}_{\rho }^{T}\mathbf{A}_{\rho }^{T}\mathbf{Y})  \\
\end{matrix} \right],
 \end{multline*}
into which substitute (\ref{W2AinvToFDecreaseNodes134}) to obtain
\begin{subnumcases}{\label{}}
\mathbf{W}_{k}^{1:(k-\rho ),:}=\mathbf{W}_{k-\rho }^{{}}+\mathbf{T}_{\rho }^{{}}(\mathbf{T}_{\rho }^{T}\mathbf{A}_{k-\rho }^{T}\mathbf{Y}+\mathbf{G}_{\rho }^{T}\mathbf{A}_{\rho }^{T}\mathbf{Y}) &  \label{wWaveWTTppaa}\\
\mathbf{W}_{k}^{(k-\rho +1):k,:}=\mathbf{G}_{\rho }^{{}}(\mathbf{T}_{\rho }^{T}\mathbf{A}_{k-\rho }^{T}\mathbf{Y}+\mathbf{G}_{\rho }^{T}\mathbf{A}_{\rho }^{T}\mathbf{Y})  &  \label{wWaveWTTppbb}
\end{subnumcases}
%$\mathbf{W}_{k}^{1:(k-\rho ),:}=\mathbf{W}_{k-\rho }^{{}}+\mathbf{T}_{\rho }^{{}}(\mathbf{T}_{\rho }^{T}\mathbf{A}_{k-\rho }^{T}\mathbf{Y}+\mathbf{G}_{\rho }^{T}\mathbf{A}_{\rho }^{T}\mathbf{Y})$   (wWaveWTTppaa)
%$\mathbf{W}_{k}^{(k-\rho +1):k,:}=\mathbf{G}_{\rho }^{{}}(\mathbf{T}_{\rho }^{T}\mathbf{A}_{k-\rho }^{T}\mathbf{Y}+\mathbf{G}_{\rho }^{T}\mathbf{A}_{\rho }^{T}\mathbf{Y})$   (wWaveWTTppbb)
%where ${\bf{W}}_k^{i:j,:}$ denotes the sub-matrix in ${\bf{W}}_k$ from the $i^{th}$ row to the $j^{th}$ row.
From  (\ref{wWaveWTTppbb})  we deduce $\mathbf{T}_{\rho }^{T}\mathbf{A}_{k-\rho }^{T}\mathbf{Y}+\mathbf{G}_{\rho }^{T}\mathbf{A}_{\rho }^{T}\mathbf{Y}=\mathbf{G}_{\rho }^{-1}\mathbf{W}_{k}^{(k-\rho +1):k,:}$, which is then substituted into (\ref{wWaveWTTppaa}) to obtain
(\ref{Wkrho2wkTGW834393}).

\section{Derivation of   (\ref{Lwave2IKK40425bothDecrease}),  (\ref{Qldelta2832DecreaseFinal}),
(\ref{Bwave2832DecreaseFinal}),
(\ref{xWmnIncreaseCompute86759DecreaseFinal}) and (\ref{xWmnIncreaseCompute4L9345Decrease})}
%(\ref{Lwave2IKK40425bothDecrease}) and
%Now we need to compute
%\begin{equation}\label{Qm1AAIdefine23213Decrease}
%{{\mathbf{Q}}_{\bar l - \bar \delta}}={{\left( {{\bf{A}}_{\bar l - \bar \delta}^{T}}{\bf{A}}_{\bar l - \bar \delta}+\lambda \mathbf{I} \right)}^{-1}}.
%\end{equation}%[Qm1AAIdefine23213]
From (\ref{AxInputDecrease31232}) and
(\ref{Qm1AAIdefine23213}),
%(\ref{Qm1AAIdefine23213Decrease}),
we can deduce
\begin{equation}\label{Q2AAIAtAinv4133467Decrease}
{{\mathbf{Q}}_{\bar l - \bar \delta}}={{\left( \left( {{\bf{A}}_{\bar l}^{T}}{\bf{A}}_{\bar l}+\lambda \mathbf{I} \right)-{\bf{A}}_{\bar \delta}^{T}{\bf{A}}_{\bar \delta} \right)}^{-1}}.
\end{equation}%[Q2AAIAtAinv4133467]
%By comparing
%It can easily be seen that
Obviously we can replace ${{\mathbf{Q}}_{\bar l}}$,  ${{\bf{A}}_{\bar l - \bar p}^{T}}$
  ${\bf{A}}_{\bar l - \bar p}$,
 $\mathbf{A}_{x}^{T}$ and $-\mathbf{A}_{x}$ in (\ref{Q2AAIAtAinv4133467}) with ${{\mathbf{Q}}_{\bar l - \bar \delta}}$,
 ${{\bf{A}}_{\bar l}^{T}}$, ${\bf{A}}_{\bar l}$, ${\bf{A}}_{\bar \delta}^{T}$ and ${\bf{A}}_{\bar \delta}$, respectively, to obtain
 (\ref{Q2AAIAtAinv4133467Decrease}), and then we can make the same replacement in (\ref{Lwave2IKK40425both}) and  (\ref{yyyyxxxxx14223}), to obtain
     (\ref{Lwave2IKK40425bothDecrease}) and
  \begin{subnumcases}{\label{yyyyxxxxx14223DecreaseTemp}}
\delta \le k: \left\{ \begin{array}{l}
{{\mathbf{B}}}= {{\mathbf{Q}}_{\bar l} }{\bf{A}}_{\bar \delta}^{T}{{(\mathbf{I}-{\bf{A}}_{\bar \delta}^{{}}{{\mathbf{Q}}_{\bar l} }{\bf{A}}_{\bar \delta}^{T})}^{-1}} \\
{{\mathbf{Q}}_{\bar l - \bar \delta}}={{\mathbf{Q}}_{\bar l} }+{{\mathbf{B}}} {\bf{A}}_{\bar \delta}^{{}}{{\mathbf{Q}}_{\bar l} }
\end{array} \right.,  &  \label{DtAaaa1234yyyyxxDecreaseTemp}\\
\delta \ge k: \left\{ \begin{array}{l}
{{\mathbf{Q}}_{\bar l - \bar \delta}}={{(\mathbf{I}-{{\mathbf{Q}}_{\bar l} }{\bf{A}}_{\bar \delta}^{T} {\bf{A}}_{\bar \delta}^{{}})}^{-1}}{{\mathbf{Q}}_{\bar l} } \\
{{\mathbf{B}}}= {{\mathbf{Q}}_{\bar l - \bar \delta}} {\bf{A}}_{\bar \delta}^{T}
\end{array} \right.,  & \label{CAxbbb14142yyyyxxDecreaseTemp}
\end{subnumcases}
respectively.
From (\ref{yyyyxxxxx14223DecreaseTemp}) we can deduce  (\ref{Qldelta2832DecreaseFinal}) and (\ref{Bwave2832DecreaseFinal}).
%We can write (\ref{yyyyxxxxx14223DecreaseTemp}) as

%We need to write or deduce
%\begin{equation}\label{xWbarMN2AbarYYa1341Whole}
%{\mathbf{W}}_{\bar l}={{\mathbf{Q}}_{\bar l}}{{\bf{A}}_{\bar l}^{T}} {{\mathbf{Y}}_{\bar l}}.
% \end{equation}%[]
%\subsection{Ridge Solution by Inversion of the Hermitian Matrix in the Ridge Inverse of the Row-Partitioned Matrix}
From  (\ref{xWbarMN2AbarYYa1341})
%(\ref{xWbarMN2AbarYYa1341Whole})
 we can obtain
%\begin{equation}\label{xWbarWholeldelta112}
${\mathbf{W}}_{\bar l-\bar \delta}={{\mathbf{Q}}_{\bar l -\bar \delta}}{{\bf{A}}_{\bar l-\bar \delta}^{T}} {{\mathbf{Y}}_{\bar l-\bar \delta}}$,
%\end{equation}
into which we substitute (\ref{Qldelta2832DecreaseFinal}) to obtain
\begin{equation}\label{xWbarWholeldelta112}
{\mathbf{W}}_{\bar l-\bar \delta}= {{\mathbf{\tilde B}}}{{\mathbf{Q}}_{\bar l} }{{\bf{A}}_{\bar l-\bar \delta}^{T}} {{\mathbf{Y}}_{\bar l-\bar \delta}}.
\end{equation}
On the other hand, let us
substitute (\ref{AxInputDecrease31232})
% into (\ref{xA2QAt13413}), and then substitute (\ref{xA2QAt13413})
 and
 (\ref{Ypermuted7391742})  into (\ref{xWbarMN2AbarYYa1341})
to obtain ${\mathbf{W}}_{\bar l}={{\mathbf{Q}}_{\bar l}} \left[ \begin{array}{l}
{\bf{A}}_{\bar l - \bar \delta}\\
{\bf{A}}_{\bar \delta}
\end{array} \right]^{T} \left[ \begin{matrix}
   {{\mathbf{Y}}_{\bar l - \bar \delta}}  \\
   {{\mathbf{Y}}_{\bar \delta}}  \\
\end{matrix} \right]$, i.e.,
%\begin{equation}\label{xWbarMN2AbarYYa1341Decrease}
%{\mathbf{W}}_{\bar l}={{\mathbf{Q}}_{\bar l}}{\bf{A}}_{\bar l - \bar \delta}^{T}{{\mathbf{Y}}_{\bar l - \bar \delta}}
%+{{\mathbf{Q}}_{\bar l}}{\bf{A}}_{\bar \delta}^{T}{{\mathbf{Y}}_{\bar \delta}}.
% \end{equation}%[]
 \begin{equation}\label{xWbarDecreaseUse}
{{\mathbf{Q}}_{\bar l}}{\bf{A}}_{\bar l - \bar \delta}^{T}{{\mathbf{Y}}_{\bar l - \bar \delta}}=
{\mathbf{W}}_{\bar l}-{{\mathbf{Q}}_{\bar l}}{\bf{A}}_{\bar \delta}^{T}{{\mathbf{Y}}_{\bar \delta}}.
 \end{equation}%[]
Finally we can substitute (\ref{xWbarDecreaseUse}) into (\ref{xWbarWholeldelta112}) to obtain (\ref{xWmnIncreaseCompute86759DecreaseFinal}).
%${\mathbf{W}}_{\bar l-\bar \delta}= {{\mathbf{\tilde B}}}({\mathbf{W}}_{\bar l}-{{\mathbf{Q}}_{\bar l}}{\bf{A}}_{\bar \delta}^{T}{{\mathbf{Y}}_{\bar \delta}})$,

Substitute (\ref{Lbig2LLwave59056Decrease}) into (\ref{Q2PiPiT9686954}) to obtain ${{\mathbf{Q}}_{\bar l -\bar \delta}}={{\mathbf{F}}_{\bar l - \bar \delta}} \mathbf{V}^T {{\mathbf{F}}_{\bar l}^T }$,  which is then substituted
%and then substitute (\ref{Q2PiPiT9686954})
 into
(\ref{xWbarWholeldelta112}) to obtain
\begin{equation}\label{xWbarWholeldelta112C1}
{\mathbf{W}}_{\bar l-\bar \delta}={{\mathbf{F}}_{\bar l - \bar \delta}} \mathbf{V}^T {{\mathbf{F}}_{\bar l}^T } {{\bf{A}}_{\bar l-\bar \delta}^{T}} {{\mathbf{Y}}_{\bar l-\bar \delta}},
\end{equation}
i.e.,
\begin{equation}\label{xWbarWholeldelta112C2}
{\mathbf{W}}_{\bar l-\bar \delta}={{\mathbf{F}}_{\bar l - \bar \delta}} \mathbf{V}^T  {{\mathbf{F}}_{\bar l}^{-1} } {{\mathbf{F}}_{\bar l} } {{\mathbf{F}}_{\bar l}^T } {{\bf{A}}_{\bar l-\bar \delta}^{T}} {{\mathbf{Y}}_{\bar l-\bar \delta}}.
\end{equation}
Then
%into which
 substitute (\ref{Q2PiPiT9686954}) into (\ref{xWbarWholeldelta112C2}) to obtain
\begin{equation}\label{xWbarWholeldelta112C3}
{\mathbf{W}}_{\bar l-\bar \delta}={{\mathbf{F}}_{\bar l - \bar \delta}} \mathbf{V}^T  {{\mathbf{F}}_{\bar l}^{-1} } {{\mathbf{Q}}_{\bar l - \bar \delta}} {{\bf{A}}_{\bar l-\bar \delta}^{T}} {{\mathbf{Y}}_{\bar l-\bar \delta}},
\end{equation}
into which substitute (\ref{xWbarDecreaseUse}) to obtain
\begin{equation}\label{xWbarWholeldelta112C4}
{\mathbf{W}}_{\bar l-\bar \delta}={{\mathbf{F}}_{\bar l - \bar \delta}} \mathbf{V}^T  {{\mathbf{F}}_{\bar l}^{-1} } ({\mathbf{W}}_{\bar l}-{{\mathbf{Q}}_{\bar l}}{\bf{A}}_{\bar \delta}^{T}{{\mathbf{Y}}_{\bar \delta}}).
\end{equation}
Finally let us
%into which
 substitute (\ref{Q2PiPiT9686954}) into (\ref{xWbarWholeldelta112C4}) to obtain

${\mathbf{W}}_{\bar l-\bar \delta}={{\mathbf{F}}_{\bar l - \bar \delta}} \mathbf{V}^T  {{\mathbf{F}}_{\bar l}^{-1} } ({\mathbf{W}}_{\bar l}-{{\mathbf{F}}_{\bar l} } {{\mathbf{F}}_{\bar l}^{T} }  {\bf{A}}_{\bar \delta}^{T}{{\mathbf{Y}}_{\bar \delta}})$, i.e.,
    \begin{equation}\label{}
{\mathbf{W}}_{\bar l-\bar \delta}= {{\mathbf{F}}_{\bar l - \bar \delta}} \mathbf{V}^T  {{\mathbf{F}}_{\bar l}^{-1} }{\mathbf{W}}_{\bar l}-{{\mathbf{F}}_{\bar l - \bar \delta}} \mathbf{V}^T  {{\mathbf{F}}_{\bar l}^{T} }  {\bf{A}}_{\bar \delta}^{T}{{\mathbf{Y}}_{\bar \delta}},
 \end{equation}
into which we can substitute (\ref{Lbig2LLwave59056Decrease}) to obtain (\ref{xWmnIncreaseCompute4L9345Decrease}).

\ifCLASSOPTIONcaptionsoff
  \newpage
\fi

\end{document}